\documentclass[journal,twoside]{IEEEtran}
\usepackage[noadjust,compress]{cite}
\usepackage[hidelinks]{hyperref}
\usepackage[cmex10]{amsmath}
\usepackage{amsfonts}
\usepackage{epsfig}
\usepackage{algorithm}
\usepackage{algorithmic}
\usepackage{balance}
\usepackage{color}
\usepackage[caption=false,font=footnotesize]{subfig}
\usepackage{rotating}
\usepackage{multirow}

\allowdisplaybreaks[4]

\graphicspath{{figures/}}

\newtheorem{lemma}{Lemma}

\newcommand{\Prob}[1]{\text{Prob}\left\{#1\right\}}
\newcommand{\E}[1]{\mathbb{E}\left[#1\right]}
\DeclareMathOperator*{\argmax}{arg\,max}
\DeclareMathOperator*{\argmin}{arg\,min}
\newcommand{\ie}{\emph{i}.\emph{e}.}
\newcommand{\eg}{\emph{e}.\emph{g}.}
\newcommand{\etal}{\emph{et al}.}
\newcommand{\cf}{\emph{cf}.}

\newlength{\nomenVar}
\newlength{\nomenDes}
\title{Enhancing Electricity-System Resilience with Adaptive Robust Optimization and Conformal Uncertainty Characterization}
\author{
	Shuyi~Chen,~\IEEEmembership{Student~Member,~IEEE,}
	Shixiang~Zhu,~\IEEEmembership{Member,~IEEE,}
	and~Ramteen~Sioshansi,~\IEEEmembership{Fellow,~IEEE}%
	\thanks{Manuscript received $30$~September,~$2025$. 
	This work was supported by Electric Power Research Institute grant~DKT$230520$, Carnegie Mellon Electricity Industry Center, and Wilton E.\ Scott Institute for Energy Innovation. \textit{(Corresponding author: Shuyi Chen.)}}
	\thanks{S.\ Chen and S.\ Zhu are with Heinz College of Information Systems and Public Policy, Carnegie Mellon University, Pittsburgh, PA $15213$-$3815$, USA (e-mail: \href{mailto:shuyic@alumni.cmu.edu}{shuyic@alumni.cmu.edu} and \href{mailto:shixianz@andrew.cmu.edu}{shixianz@andrew.cmu.edu}).}
	\thanks{R.\ Sioshansi is with Department of Engineering and Public Policy, Carnegie Mellon Electricity Industry Center, Department of Electrical and Computer Engineering, Heinz College of Information Systems and Public Policy, and Wilton E.\ Scott Institute for Energy Innovation, Carnegie Mellon University, Pittsburgh, PA $15213$-$3815$, USA and Department of Integrated Systems Engineering, The Ohio State University, Columbus, OH $43210$-$1271$, USA (e-mail: \href{mailto:rsioshan@andrew.cmu.edu}{rsioshan@andrew.cmu.edu}).}
}

\begin{document}
\maketitle
\bibliographystyle{IEEEtran}

\begin{abstract}
Extreme weather is straining electricity systems, exposing the limitations of reactive responses, and prompting the need for proactive resilience planning. Most existing approaches to enhance electricity-system resilience employ simplified uncertainty models and decouple proactive and reactive decisions. This paper proposes a novel tri-level optimization model that integrates proactive actions, adversarial disruptions, and reactive responses. Conformal prediction is used to construct distribution-free system-disruption uncertainty sets with coverage guarantees. The tri-level problem is solved by using duality theory to derive a bi-level reformulation and employing Benders's decomposition. Numerical experiments demonstrate that our approach outperforms conventional robust and two-stage methods.
\end{abstract}


\section*{Nomenclature}
\subsection*{Optimization-Model Sets and Parameters}
\noindent
\begin{tabular}{@{}p{\nomenVar}p{\nomenDes}@{}}
	$B$&			proactive-action budget (\$)\\
	$b_i$&			region-$i$ proactive-action cost (\$)\\
	$\mathcal{B}(B)$&	proactive-action feasible region\\
	$C$&			reactive-action budget (\$)\\
	$c_i$&			region-$i$ reactive-action cost (\$)\\
	$\mathcal{C}(C)$&	reactive-action feasible region\\
	$h_i$&			region-$i$ per-unit outage cost (\$)\\
	$\hat{L}_0(w)$&		lower-bound of global confidence interval\\
	$\hat{L}_i(w)$&		lower-bound of region-$i$ confidence interval\\
	$n$&			number of regions in the electricity system\\
	$\hat{T}_0(w)$&		upper-bound of global confidence interval\\
	$\hat{T}_i(w)$&		upper-bound of region-$i$ confidence interval\\
	$\hat{u}_i$&		forecast of the number of region-$i$ outages\\
	$\bar{u}_i$&		realized number of region-$i$ outages\\
	$w$&			contextual-feature matrix\\
	$\bar{x}_i$&		fixed region-$i$ proactive action\\
	$\bar{y}_i$&		fixed region-$i$ reactive action\\
	$\alpha$&		miscoverage level of uncertainty set
\end{tabular}

\noindent
\begin{tabular}{@{}p{\nomenVar}p{\nomenDes}@{}}
	$\Omega(w)$&		uncertainty set for $u$
\end{tabular}

\subsection*{Optimization-Model Decision Variables}
\noindent
\begin{tabular}{@{}p{\nomenVar}p{\nomenDes}@{}}
	$u_i$&	number of region-$i$ outages\\
	$x_i$&	equals~$1$ if region~$i$ receives proactive action, equals~$0$ otherwise\\
	$y_i$&	equals~$1$ if region~$i$ receives reactive action, equals~$0$ otherwise
\end{tabular}

\subsection*{Conformal-Prediction Sets and Models}
\noindent
\begin{tabular}{@{}p{\nomenVar}p{\nomenDes}@{}}
	$\hat{f}(\cdot)$&		predictive model that estimates $\E{u \vert w}$\\
	$\mathcal{M}$&			set of historical observations of $w$ and $u$\\
	$\mathcal{M}_\textrm{cal}$&	calibration set of historical observations\\
	$\mathcal{M}_\textrm{train}$&	training set of historical observations
\end{tabular}

\subsection*{Benders's Decomposition Parameters}
\noindent
\begin{tabular}{@{}p{\nomenVar}p{\nomenDes}@{}}
	$\mathcal{T}$&	iteration limit\\
	$\epsilon$&	convergence tolerance\\
	$\Phi^+$&	objective-function upper bound\\
	$\Phi^-$&	objective-function lower bound
\end{tabular}

\section{Introduction}
\IEEEPARstart{E}{xtreme} weather is becoming more frequent and severe and placing increasing stress on electricity systems \cite{StankovicTomsovic2018,HeHuangLiuZhang2021}. For example, during March~$2018$, the northeastern United States of America (US) experienced winter storms causing power failures affecting more than $2\,755\,000$~customers across New England and \$$4$~billion of economic losses.\footnote{\href{https://tinyurl.com/6tmjuvn2}{https://tinyurl.com/6tmjuvn2}} Such events underscore electricity-system vulnerability and the need to increase their resilience \cite{LiuShahidehpourLiLiuCaoBie2017,PanteliTrakasMancarellaHatziargyriou2017,StankovicTomsovic2018,StankovicTomsovicDeCaroBraunChowCukalevski2023}.

Most works that examine electricity-system resilience focus on reactive strategies (\eg, dispatching repair crews, deploying mobile supply units, and rerouting electricity through alternative paths) that are taken during or after an event \cite{HuangWangChenQiGuo2017,LinChenWangBie2019,LeiChenLiHou2019}. However, real-time conditions (\eg, downed transmission or distribution lines) may limit the effectiveness of such actions \cite{MohamedChenSuJin2019}. Consequently, there is need for proactive actions (\eg, hardening or undergrounding infrastructure, managing vegetation, or pre-positioning crews or mat\'{e}riel) that are taken before an event \cite{MaChenWang2018,HuangWangChenQiGuo2017,BaoZhangOuyangMiao2019,MohamedChenSuJin2019,WangRousisStrbac2021,HinesDobsonRezaei2017}.

Electricity-system resilience is enhanced most effectively if forecasting (of events and their severity) informs the co-optimization of reactive and proactive strategies. Most methods that are proposed in the literature do not meet this goal. Many methods use heuristic or simplified uncertainty representations (\eg, fixed probability distributions, historical averages, or hand-crafted scenarios) that do not capture high-dimensional, spatio-temporal complexity of weather-driven disruptions and lack formal predictive guarantees. These methods can over-fit to past data or under-estimate plausible extreme scenarios. Moreover, many methods treat proactive and reactive decisions in isolation (\eg, by omitting reactive decisions or oversimplifying them). In reality, such decisions should be coupled tightly---proactive actions determine feasible reactive actions and reactive capacity should influence where proactive actions are taken. Ignoring this interdependence can yield poor decisions.

Addressing these limitations of the extant literature introduces three core methodological challenges, which this paper addresses. First, the spatio-temporal impacts of extreme events vary drastically, making such events non-stationary, high-dimensional, and poorly suited for traditional parametric or distribution-based models \cite{VovkGammermanShafer2022,ZengZhao2013,PatelRayanTewari2024}. Second, extreme events are rare and regionally imbalanced, with limited, noisy, and sparse historical records. These properties render standard statistical learning unreliable and call for distribution-free or data-efficient methods. Finally, interactions between uncertain events and proactive and reactive decisions yields a nested decision problem. Unlike standard robust optimization, the objective of this setting depends on decisions that are made before and after uncertainty is realized.

We address these challenges by developing a novel tri-level robust-optimization model that captures explicitly interactions between proactive and reactive decisions and uncertainty \cite{BertsimasBrownCaramanis2011}. The top level determines budget-constrained proactive decisions. The middle level selects an adversarial outage scenario that is drawn from a high-probability uncertainty set that is constructed from spatio-temporal conformal prediction \cite{PatelRayanTewari2024}. The lower level models budget-constrained reactive decisions. We employ duality theory to reformulate the tri-level model as an equivalent bi-level program, to which we apply Benders's decomposition \cite{Benders1962}. We validate and demonstrate our approach through a numerical example and a real-world case study. Our results reveal three insights. Optimizing proactive and reactive decisions jointly yields lower worst-case losses compared to treating such decisions separately. Second, conformal-prediction-based uncertainty sets enhance robustness by adapting to spatial heterogeneity. Finally, the model scales well.

There are works in the literature that apply robust optimization to model infrastructure-hardening, network-investment, and other electricity-system problems \cite{Jabr2013}. Robust optimization provides solutions that protect against worst-case outcomes, meaning that it can be conservative \cite{Ben-TalGoryashkoGuslitzerNemirovski2004,BertsimasNaStellatoWang2025}. Adaptive robust optimization can mitigate this conservatism by representing recourse decisions that react to uncertain outcomes \cite{BertsimasLitvinovSunZhaoZheng2013}. One standard approach to solving (adaptive) robust optimization problems is column-and-constraint generation \cite{ZengZhao2013,Zeng2020,ZengDongSioshansiXuZeng2020}. Our model differs from this existing literature in two ways. First, unlike standard methods, our approach models explicitly sequential interactions between proactive decisions, realized disruptions, and reactive responses. Second, for tractability, most adaptive-robust-optimization models assume an objective function with an additively separable structure. Our model generalizes this assumption by taking an unified approach to modeling proactive and reactive decisions.

As an alternative to robust optimization, one could employ decision-dependent uncertainty, which captures explicitly the effect of proactive decisions on the structure of uncertainties that are related to system disruptions \cite{NohadaniSharma2018}. Our model differs from such approaches by accounting for endogeneity of uncertainty and by allowing downstream adaptivity and budgeted recourse actions, which is a more realistic decision-planning loop. Another approach to modeling electricity-system resilience is to employ defender-attacker-defender models, which represent a defender's investment, an attacker's disruption, and a subsequent defensive response \cite{BrownCarlyleSalmeronWood2006}. This approach is applied to electricity-system-resilience modeling, and assumes typically that hardened assets are removed from the attacker's feasible set, thereby simplifying interactions between proactive and reactive decisions \cite{HuangWangChenQiGuo2017,MaChenWang2018,BaoZhangOuyangMiao2019,WangRousisStrbac2021}. Another common simplification is to assume combinatorial or polyhedral uncertainty sets with deterministic attack budgets or spatial limits \cite{MaChenWang2018,BaoZhangOuyangMiao2019}. Our approach differs from these in two key respects. First, our focus is natural hazards rather than a strategic adversary, which allows protected assets to be affected. Second, we construct localized, spatio-temporal uncertainty regions using conformal prediction, which provides finite-sample coverage guarantees and heterogeneous, region-specific risk bounds that enhance the reliability of uncertainty-set estimation compared to traditional methods.

Modeling electricity-system disruptions is an active area of research, which employs numerous techniques including machine learning, Poisson processes, neural networks, and graph-based models \cite{HinesDobsonRezaei2017,EskandarpourKhodaei2017}. Prediction-focused methods tend to rely on assumptions, such as stationarity or parametric distributions, which may be unrealistic, especially under rare and extreme conditions \cite{LiWangGuoZhang2024}. Conformal prediction is an alternative technique, which provides distribution-free prediction sets with finite-sample uncertainty-coverage guarantees \cite{VovkGammermanShafer2022}. Non-stationarity (\eg, due to extreme weather) can be addressed with graph-based conformal prediction. Standard conformal prediction assumes homoskedastic noise, but normalized conformal prediction can relax this assumption by scaling residuals with local variability \cite{PapadopoulosGammermanVovk2008,Bellotti2020}. Building on these advances, Patel \etal\ \cite{PatelRayanTewari2024} propose a framework that integrates normalized conformal prediction with robust optimization to support distribution-free, context-aware decision-making. Our work builds on these advances by embedding normalized conformal prediction into the middle level of our optimization, which enables planning under heteroskedastic, weather-driven uncertainty. The distribution-free nature of conformal prediction ensures that the model is valid even if the underlying distributions are unknown or non-stationary \cite{PatelRayanTewari2024}.

This paper makes four major contributions to the extant literature. First, we develop a tri-level model that optimizes proactive and reactive actions jointly and under uncertainty. Second, we develop a spatio-temporal conformal-prediction method to construct data-driven and distribution-free uncertainty sets of outages with uncertainty-coverage guarantees. Third, we propose an algorithm with which to solve the tri-level model efficiently. Finally, we validate and demonstrate the proposed model through numerical experiments.

The remainder of this paper is organized as follows. Section~\ref{sec:model} details the tri-level model and how we assess its performance. Section~\ref{sec:solution} describes our solution method. Sections~\ref{sec:example} and~\ref{sec:caseStudy} summarize our example and case study, respectively. Section~\ref{sec:concl} concludes.

\section{Model Formulation}
\label{sec:model}
We detail our model by providing its formulation in Section~\ref{sec:model:optimization} and describing in Section~\ref{sec:model:predict} the use of conformal prediction to generate the uncertainty set for its middle level. Section~\ref{sec:model:benchmark} summarizes how we assess model performance.

\subsection{Tri-Level Model}
\label{sec:model:optimization}
The model is formulated as:
\begin{equation}
\label{equ:model:optimization:objFunc}
	\min_{x \in \mathcal{B}(B)} \max_{u \in \mathcal{U}(w,\alpha)} \min_{y \in \mathcal{C}(C)} \sum_{i=1}^n h_i u_i \cdot (1-x_i) (1-y_i);
\end{equation}
where:
\begin{equation}
\label{equ:model:optimization:cons:proactive}
	\mathcal{B}(B) = \left\{
		x \in \{0,1\}^n \left\vert
			\sum_{i=1}^n b_i x_i \leq B
		\right.
	\right\};
\end{equation}
\begin{equation}
\label{equ:model:optimization:cons:nature}
	\mathcal{U}(w,\alpha) = \{
		\Prob{u \in \Omega(w)} \geq 1-\alpha
	\};
\end{equation}
and:
\begin{equation}
\label{equ:model:optimization:cons:reactive}
	\mathcal{C}(C) = \left\{
		y \in \{0,1\}^n \left|
			\sum_{i=1}^n c_i y_i \leq C
		\right.
	\right\}.
\end{equation}
The objective that is in~(\ref{equ:model:optimization:objFunc}) is total outage costs. For all $i = 1,\dots,n$, $h_i$ is region~$i$'s (a region could be, \eg, a geographic area or distribution circuit) per-outage cost and $u_i$ is the number of region-$i$ outages. Thus, $\forall i = 1,\dots,n$, the product, $h_i u_i$, is the total cost of region-$i$ outages. The $(1-x_i) (1-y_i)$ terms that are in~(\ref{equ:model:optimization:objFunc}) are logical switches---$\forall i = 1,\dots,n$, region~$i$ has outages only if neither proactive nor reactive actions are taken (\ie, if $x_i = 0$ and $y_i = 0$). Problem~(\ref{equ:model:optimization:objFunc}) has a tri-level structure---a planner (\eg, the distribution-system owner) takes proactive and reactive actions, $x$ and $y$, respectively, to minimize outage cost, whereas nature determines outages to maximize cost. The structure whereby nature selects a worst-case $u$ and the planner selects a reactive $y$ makes the model an adaptive robust optimization.

Definitions~(\ref{equ:model:optimization:cons:proactive}) and~(\ref{equ:model:optimization:cons:reactive}) impose integrality restrictions and budget constraints on the proactive and reactive actions, respectively. $\mathcal{U}(w,\alpha)$ is defined by~(\ref{equ:model:optimization:cons:nature}) as a chance constraint, whereby $u$ must be an element of uncertainty set, $\Omega(w)$, with probability at least $1 - \alpha$. $\Omega(w)$ depends on an $n \times p$~matrix, $w$, of $p$~contextual features (\eg, historical outages, weather conditions, and infrastructure characteristics) for each region.

\subsection[Constructing Omega(w) via Conformal Prediction]{Constructing $\Omega(w)$ via Conformal Prediction}
\label{sec:model:predict}
Using an exchangeability assumption, we employ conformal prediction to construct $\Omega(w)$ \cite{PapadopoulosGammermanVovk2008,Bellotti2020,PatelRayanTewari2024}. Given a calibration set, conformal prediction ensures that for any new instance, $w$, the true response, $u$, lies within $\Omega(w)$ with probability at least $1 - \alpha$, regardless of $u$'s distribution.

To capture region-specific variability and system-wide uncertainty, we decompose $\Omega(w)$ into two components. The first requires that $u_i$ belong to the confidence interval:
\begin{equation}
\label{equ:model:predict:local}
	\Prob{\hat{L}_i(w) \leq u_i \leq \hat{T}_i(w)} \geq 1 - \alpha;	\forall i = 1,\dots,n;
\end{equation}
and the second requires that total outages to belong to the global confidence interval:
\begin{equation}
\label{equ:model:predict:global}
	\Prob{\hat{L}_0(w) \leq \sum_{i=1}^n u_i \leq \hat{T}_0(w)} \geq 1 - \alpha;
\end{equation}
with probabilities at least $1 - \alpha$. Thus, we define $\Omega(w)$ as:
\begin{multline}
\label{equ:model:predict:cons:nature}
	\Omega(w) = \left\{
		u \in \{0,1\}^n \left|
			\hat{L}_i(w) \leq u_i \leq \hat{T}_i(w);
		\vphantom{\sum_{i=1}^n}\right.
	\right.\\
	\left.
		\forall i = 1,\dots,n; \hat{L}_0(w) \leq \sum_{i=1}^n u_i \leq \hat{T}_0(w)
	\right\}.
\end{multline}

Algorithm~\ref{alg:model:predict} summarizes the steps that are used to compute $\{\hat{L}_i(w), \hat{T}_i(w)\}_{i=0}^n$. The algorithm has two inputs---$\alpha$ and $\mathcal{M}$ (\cf\ Line~\ref{alg:model:predict:input})---the latter of which is partitioned into training and calibration sets (\cf\ Line~\ref{alg:model:predict:partition}). Next, $\mathcal{M}_\mathrm{train}$ is used to train $\hat{f}(\cdot)$ (\cf\ Line~\ref{alg:model:predict:train}), which is used to generate predictions that correspond to $\{w^j\}_{j \in \mathcal{M}_\mathrm{cal}}$ (\cf\ Line~\ref{alg:model:predict:predict}).

\begin{algorithm}
\caption{Construction of $\{\hat{L}_i(w), \hat{T}_i(w)\}_{i=0}^n$}
\label{alg:model:predict}
\begin{algorithmic}[1]
	\STATE \textbf{input:} $\alpha$; $\mathcal{D} := \left\{\left(w^j,u^j\right)\right\}_{j \in \mathcal{M}}$
\label{alg:model:predict:input}
	\STATE \textbf{initialize:} $\mathcal{S}_i \gets \emptyset$, $\forall i = 0,\dots,n$
\label{alg:model:predict:init}
	\STATE divide $\mathcal{M}$ into $\mathcal{M}_\mathrm{train}$ and $\mathcal{M}_\mathrm{cal}$ with $\mathcal{M} = \mathcal{M}_\mathrm{train} \cup \mathcal{M}_\mathrm{cal}$ and $\mathcal{M}_\mathrm{train} \cap \mathcal{M}_\mathrm{cal} = \emptyset$
\label{alg:model:predict:partition}
	\STATE train $\hat{f}(w)$ on $\left\{\left(w^j,u^j\right)\right\}_{j \in \mathcal{M}_\mathrm{train}}$
\label{alg:model:predict:train}
	\STATE $\kappa^j \gets \hat{f}(w^j)$, $\forall j \in \mathcal{M}_\mathrm{cal}$
\label{alg:model:predict:predict}
	\FOR{$i \gets 0,\dots,n$}
\label{alg:model:predict:region:start}
		\IF{$i \not= 0$}
\label{alg:model:predict:nonConform:start}
			\STATE $s^j_i \gets \left\lvert\left\lvert u^j_i - \kappa^j_i\right\rvert\right\rvert$
\label{alg:model:predict:nonConform:local}
		\ELSE
\label{alg:model:predict:nonConform:else}
			\STATE $s_i^j \gets \left\lvert\left\lvert \sum\limits_{\iota=1}^n \left(u_\iota^j - \kappa^j_\iota\right)\right\rvert\right\rvert$
\label{alg:model:predict:nonConform:global}
		\ENDIF
\label{alg:model:predict:nonConform:end}
		\STATE $\mathcal{S}_i \gets \mathcal{S}_i \cup \left\{s_i^j\right\}$
\label{alg:model:predict:nonConform:classify}
		\STATE $Q_i \gets (1-\alpha)$-quantile of $S_i$
\label{alg:model:predict:quantile}
		\STATE $\hat{L}_i(w) \gets \hat{f}_i(w) - Q_i$
\label{alg:model:predict:bound:lower}
		\STATE $\hat{T}_i(w) \gets \hat{f}_i(w) + Q_i$
\label{alg:model:predict:bound:upper}
	\ENDFOR
\label{alg:model:predict:region:end}
\end{algorithmic}
\end{algorithm}

Next, Lines~\ref{alg:model:predict:region:start}--\ref{alg:model:predict:region:end} use the predictions that are generated in Line~\ref{alg:model:predict:predict} to compute non-conformity scores for $\mathcal{M}_\mathrm{cal}$ and $\{\hat{L}_i(w), \hat{T}_i(w)\}_{i=0}^n$. Specifically, $\forall i = 0,\dots,n$, Lines~\ref{alg:model:predict:nonConform:start}--\ref{alg:model:predict:nonConform:end} compute non-conformity scores between each prediction and observation that is in $\mathcal{M}_\mathrm{cal}$. If $i \not= 0$, the non-conformity scores are computed considering region-$i$ observations only (\cf\ Line~\ref{alg:model:predict:nonConform:local}), otherwise global non-conformity scores that consider all of the regions are computed if $i = 0$ (\cf\ Line~\ref{alg:model:predict:nonConform:global}). Then, $\forall i = 0,\dots,n$, Line~\ref{alg:model:predict:quantile} determines the $1-\alpha$ quantile of the region-$i$ non-conformity scores, which is used in Lines~\ref{alg:model:predict:bound:lower}--\ref{alg:model:predict:bound:upper} to determine $\hat{L}_i(w)$ and $\hat{T}_i(w)$, respectively.

The definition of $\Omega(w)$ in~(\ref{equ:model:predict:local})--(\ref{equ:model:predict:cons:nature}) yields computational and predictive benefits. $\Omega(w)$ is defined by~(\ref{equ:model:predict:cons:nature}) as a convex polyhedron, which makes~(\ref{equ:model:optimization:cons:nature}) a linear constraint. The true uncertainty set may be non-linear or non-convex. However, (\ref{equ:model:predict:cons:nature}) captures the essential geometry of the true uncertainty set with high fidelity and interpretability. The local bounds that are in~(\ref{equ:model:predict:local}) capture heterogeneous risk profiles across different regions that are in the model's geographic scope \cite{PatelRayanTewari2024}. The global bounds that are in~(\ref{equ:model:predict:global}) capture correlated risk structures, thereby ensuring coherence at the system level and that $\Omega(w)$ satisfies the desired $1-\alpha$ confidence level. This final property of~(\ref{equ:model:predict:cons:nature}) is formalized by the following lemma.

\begin{lemma}
\label{lem:model:predict}
The definition of $\Omega(w)$ in~(\ref{equ:model:predict:cons:nature}), its construction by Algorithm~\ref{alg:model:predict}, and the observations in $\mathcal{M}$ being exchangable ensure that chance constraint~(\ref{equ:model:optimization:cons:nature}) is satisfied, regardless of the distribution of $u$.
\end{lemma}

\begin{IEEEproof}
The results of Shafer and Vovk \cite{ShaferVovk2008} (\cf\ pp.~$382$ and Appendix~A) imply that the definitions of $\hat{L}_i(w)$ and $\hat{T}_i(w)$ in Lines~\ref{alg:model:predict:bound:lower} and~\ref{alg:model:predict:bound:upper}, respectively, of Algorithm~\ref{alg:model:predict} guarantee that~(\ref{equ:model:predict:local}) and~(\ref{equ:model:predict:global}) are satisfied with probability at least $1-\alpha$. Consequently, the definition of $\Omega(w)$ in~(\ref{equ:model:predict:cons:nature}) guarantees that~(\ref{equ:model:optimization:cons:nature}) is satisfied with probability at least $1-\alpha$.
\end{IEEEproof}

\subsection{Model Benchmarking}
\label{sec:model:benchmark}
We benchmark our proposed model to demonstrate how its tri-level structure (which co-optimizes proactive and reactive decisions) and use of conformal prediction improve solution quality \textit{vis-\`{a}-vis} simpler modeling approaches.

To this end, we compare solutions that are obtained from~(\ref{equ:model:optimization:objFunc}) to those that are obtained from what we term Proactive-Only and Co-Optimized Planning, both of which are deterministic. Proactive-Only Planning solves:
\begin{equation}
\label{equ:model:benchmark:proactive}
	\min_{x \in \mathcal{B}(B)} \sum_{i=1}^n h_i \hat{u}_i \cdot (1-x_i);
\end{equation}
whereas Co-Optimized Planning solves:
\begin{equation}
\label{equ:model:benchmark:coopt}
	\min_{x \in \mathcal{B}(B), y \in \mathcal{C}(C)} \sum_{i=1}^n h_i \hat{u}_i \cdot (1-x_i) (1-y_i).
\end{equation}
Solving~(\ref{equ:model:optimization:objFunc}), (\ref{equ:model:benchmark:proactive}), and (\ref{equ:model:benchmark:coopt}) yields proactive actions, which we denote as $x^\ast$, $x^\textrm{pro}$, and $x^\textrm{coopt}$, respectively. The value of:
\begin{equation}
\label{equ:model:benchmark:bilevel}
	\max_{u \in \bar{\mathcal{U}}} \min_{y \in \mathcal{C}(C)} \sum_{i=1}^n h_i u_i \cdot (1-\bar{x}_i) (1-y_i);
\end{equation}
where $\bar{x}$ is fixed equal to one of $x^\ast$, $x^\textrm{pro}$, or $x^\textrm{coopt}$ can be used to compare these actions. Problem~(\ref{equ:model:benchmark:bilevel}) gives the objective-function value that results from a chosen $\bar{x}$. We consider cases in which $\bar{\mathcal{U}}$ is defined by~(\ref{equ:model:predict:cons:nature}) and in which only the local and only the global conformal prediction intervals are included in $\bar{\mathcal{U}}$. Having $\bar{\mathcal{U}}$ include only local prediction intervals is the typical `worst-case' outcome that is considered in the literature, because there is no collective outage bound that captures spatial relationships. Having $\bar{\mathcal{U}}$ include only a global prediction interval represents nature distributing the uncertainty budget adversarially across regions, because $u$ is chosen to target regions where reactive action is costly.

In addition to examining~(\ref{equ:model:benchmark:bilevel}), we contrast the solutions that are given by~(\ref{equ:model:optimization:objFunc}) and~(\ref{equ:model:benchmark:coopt}) by examining the value of:
\begin{equation}
\label{equ:model:benchmark:recourse}
	\sum_{i=1}^n h_i \bar{u}_i \cdot (1-\bar{x}_i) (1-\bar{y}_i);
\end{equation}
where $\bar{x}$ and $\bar{y}$ are fixed equal to the decisions that are given by one of~(\ref{equ:model:optimization:objFunc}) or~(\ref{equ:model:benchmark:coopt}).

We examine four approaches to generating $\hat{u}$. The first---Empirical Average---has $\hat{u}_i = \eta_i$, where:
\[
	\eta_i = \frac{1}{\lvert \mathcal{M} \rvert} \sum_{j \in \mathcal{M}} u^j_i; \forall i = 1,\dots,n;
\]
or the historical sample mean of $u_i$. The second, which we call Empirical Conservative, has $\hat{u}_i = \eta_i + 1.96 \sigma_i$, where:
\[
	\sigma_i = \sqrt{\frac{1}{\lvert \mathcal{M} \rvert-1} \sum_{j \in \mathcal{M}} \left(u^j_i-\eta_i\right)^2}; \forall i = 1,\dots,n;
\]
or the upper bound of a two-sided $95$\% confidence interval of the mean of $u_i$. The third---Conformal Average---has:
\[
	\hat{u}_i = \frac{1}{\lvert \mathcal{M}_\mathrm{train} \rvert} \sum_{j \in \mathcal{M}_\mathrm{train}}\hat{f}(w^j_i); \forall i = 1,\dots,n;
\]
which is a sample mean that is generated by the prediction model. The fourth, which we term Conformal Conservative, has $\hat{u}_i = \hat{T}_i(w)$, $\forall i = 1,\dots,n$. As for~(\ref{equ:model:benchmark:recourse}), we consider cases in which $\bar{u}_i$ equals $\eta_i$, $\eta_i + \sigma_i$, and $\eta_i + 2\sigma_i$, $\forall i = 1,\dots,n$.

\section{Solution Strategy}
\label{sec:solution}
Problem~(\ref{equ:model:optimization:objFunc}) is an intractable tri-level model with a tri-linear objective function and lower-level binary variables. We employ the following steps to simplify this problem structure.

\subsection{Continuous Reformulation of Recourse Problem}
\label{sec:solution:continuous}
For fixed $x$ and $u$, (\ref{equ:model:optimization:objFunc}) is equivalent to:
\begin{equation}
\label{equ:solution:continuous:objFunc:fixed}
	\min_{y \in \mathcal{C}(C)} \sum_{i=1}^n \zeta_i \cdot (1-y_i);
\end{equation}
where $\zeta_i = h_i u_i \cdot (1-x_i)$, $\forall i = 1,\dots,n$. The following lemma shows that~(\ref{equ:solution:continuous:objFunc:fixed}) and its continuous relaxation are equivalent.

\begin{lemma}
\label{lem:solution:continuous}
Let $y^\ast$ and $\hat{y}^\ast$ be global optima of~(\ref{equ:solution:continuous:objFunc:fixed}) and:
\begin{equation}
\label{equ:solution:continuous:objFunc:relax}
	\min_{y \in \hat{\mathcal{C}}(C)} \sum_{i=1}^n \zeta_i \cdot (1-y_i);
\end{equation}
respectively, where:
\[
	\hat{\mathcal{C}}(C) = \left\{
		y \in \mathbb{R}^n \left|
			\sum_{i=1}^n c_i y_i \leq C, y_i \in [0,1], \forall i = 1,\dots,n
		\right.
	\right\}.
\]
If $C \in \mathbb{Z}_+$ and all non-zero values of $\{\zeta_i\}_{i=1}^n$ are unique, then $\hat{y}^\ast \in \{0,1\}^n$ and we have that:
\[
	\sum_{i=1}^n \zeta_i \cdot (1-y^\ast_i) = \sum_{i=1}^n \zeta_i \cdot (1-\hat{y}^\ast_i).
\]
\end{lemma}

\begin{IEEEproof}
$\hat{\mathcal{C}}(C)$, can be written as, $Ay \leq b$, where:
\[
	A = \left[\begin{array}{ccc}
		E^\top&	I_n&	-I_n
	\end{array}\right]^\top;
\]
\[
	b = \left(\begin{array}{ccc}
		C&	1&	0
	\end{array}\right)^\top;
\]
$E$ is an $n$-dimensional vector of ones, and $I_n$ is an $n \times n$ identity matrix. Due to $A$ being totally unimodular and $b$ being integral, every extreme point of $\hat{\mathcal{C}}(C)$ is integral \cite{Wolsey1998}. Moreover, because the non-zero values of $\{\zeta_i\}_{i=1}^n$ are unique, an optimal solution of~(\ref{equ:solution:continuous:objFunc:relax}) must be at an extreme point (not a polyhedral face) of $\hat{\mathcal{C}}(C)$. Consequently, we must have $\hat{y}^\ast \in \{0,1\}^n$. The final part of the lemma follows from $\hat{\mathcal{C}}(C)$ being a relaxation of $\mathcal{C}(C)$---$\hat{y}^\ast$ is feasible in $\mathcal{C}(C)$. Thus, the optimized value of~(\ref{equ:solution:continuous:objFunc:fixed}) cannot be better than that of~(\ref{equ:solution:continuous:objFunc:relax}).
\end{IEEEproof}

As a result of Lemma~\ref{lem:solution:continuous}, (\ref{equ:model:optimization:objFunc}) is equivalent to:
\begin{equation}
\label{equ:solution:continuous:objFunc}
	\min_{x \in \mathcal{B}(B)} \max_{u \in \mathcal{U}(w,\alpha)} \min_{y \in \hat{\mathcal{C}}(C)} \sum_{i=1}^n h_i u_i \cdot (1-x_i) (1-y_i);
\end{equation}
which has linear middle- and lowest-level problems.

\subsection{Combining Middle and Lowest Levels of~(\ref{equ:solution:continuous:objFunc})}
\label{sec:solution:combine}
The lowest level of~(\ref{equ:solution:continuous:objFunc}) can be written as:
\begin{align}
\label{equ:solution:combine:primal:first}
\min_y\ &
	\sum_{i=1}^n -\zeta_i y_i\\
\mathrm{s.t.}\ &
	\sum_{i=1}^n c_i y_i \leq C;&&					(\lambda)\\
\label{equ:solution:combine:primal:last}
&	0 \leq y_i \leq 1;		\forall i = 1,\dots,n;&&	(\mu_i)
\end{align}
where the dual variable that is associated with each constraint is given in parentheses to its right and the constant:
\[
	\sum_{i=1}^n \zeta_i;
\]
is excluded from~(\ref{equ:solution:combine:primal:first}). The dual of~(\ref{equ:solution:combine:primal:first})--(\ref{equ:solution:combine:primal:last}) is:
\begin{align}
\label{equ:solution:combine:dual:first}
\max_{\lambda,\mu}\ &
	C \lambda + \sum_{i=1}^n \mu_i\\
\mathrm{s.t.}\ &
	c_i \lambda + \mu_i \leq -\zeta_i;	\forall i = 1,\dots,n;\\
&	\lambda \leq 0;\\
\label{equ:solution:combine:dual:last}
&	\mu_i \leq 0;				\forall i = 1,\dots,n.
\end{align}
Because~(\ref{equ:solution:combine:primal:first})--(\ref{equ:solution:combine:primal:last}) is a linear optimization, it is guaranteed to exhibit strong duality, meaning that~(\ref{equ:solution:combine:primal:first})--(\ref{equ:solution:combine:primal:last}) and~(\ref{equ:solution:combine:dual:first})--(\ref{equ:solution:combine:dual:last}) have the same optimal objective-function value \cite{SioshansiConejo2017}. Consequently, (\ref{equ:solution:continuous:objFunc}) is equivalent to bi-level problem:
\begin{align}
\label{equ:solution:combine:final:objFunc}
\min_x \max_{u,\lambda,\mu}\ &
	\sum_{i=1}^n \left[
		h_i u_i \cdot (1-x_i) + \mu_i
	\right] + C \lambda\\
\label{equ:solution:combine:final:dualCons:struct}
\mathrm{s.t.}\ &
	c_i \lambda + \mu_i \leq -h_i u_i \cdot (1-x_i);	\forall i = 1,\dots,n;\\
\label{equ:solution:combine:final:dualCons:nonNeg:lambda}
&	\lambda \leq 0;\\
\label{equ:solution:combine:final:dualCons:nonNeg:mu}
&	\mu_i \leq 0;						\forall i = 1,\dots,n;\\
\label{equ:solution:combine:final:nature:local}
&	\hat{L}_i(w) \leq u_i \leq \hat{T}_i(w);		\forall i = 1,\dots,n;\\
\label{equ:solution:combine:final:nature:global}
&	\hat{L}_0(w) \leq \sum_{i=1}^n u_i \leq \hat{T}_0(w);\\
\label{equ:solution:combine:final:proactive:budget}
&	\sum_{i=1}^n b_i x_i \leq B;\\
\label{equ:solution:combine:final:proactive:binary}
&	x_i \in \{0,1\};					\forall i = 1,\dots,n.
\end{align}

\subsection{Benders's Decomposition of~(\ref{equ:solution:combine:final:objFunc})--(\ref{equ:solution:combine:final:proactive:binary})}
\label{sec:solution:benders}
As a final step, we apply Benders's decomposition to~(\ref{equ:solution:combine:final:objFunc})--(\ref{equ:solution:combine:final:proactive:binary}) \cite{Benders1962}. To do so, we define master problem:
\begin{align}
\label{equ:solution:benders:master:objFunc}
\min_{\theta,x}\ &
	\theta\\
\label{equ:solution:benders:master:optimality}
\mathrm{s.t.}\ &
	\theta \in \Theta(x);\\
\label{equ:solution:benders:master:origCons}
&	\textrm{(\ref{equ:solution:combine:final:proactive:budget}), (\ref{equ:solution:combine:final:proactive:binary})};
\end{align}
and sub-problem:
\begin{align}
\label{equ:solution:benders:sub:objFunc}
\Phi(\bar{x}) = \max_{u,\lambda,\mu}\ &
	\sum_{i=1}^n \left[
		h_i u_i \cdot (1-\bar{x}_i) + \mu_i
	\right] + C \lambda\\
\label{equ:solution:benders:sub:cons}
\mathrm{s.t.}\ &
	\textrm{(\ref{equ:solution:combine:final:dualCons:struct})--(\ref{equ:solution:combine:final:nature:global})}.
\end{align}
The master problem has variables $\theta$, which is the master-problem estimate of the value of~(\ref{equ:solution:combine:final:objFunc}), and $x$. Constraint~(\ref{equ:solution:benders:master:optimality}) restricts $\theta$ using a collection, $\Theta(x)$, of optimality cuts (which depend on the choice of $x$) that are separated using the sub-problem. Constraints~(\ref{equ:solution:benders:master:origCons}) impose the budget and integrality restrictions on $x$ that are in~(\ref{equ:solution:combine:final:objFunc})--(\ref{equ:solution:combine:final:proactive:binary}). Sub-problem~(\ref{equ:solution:benders:sub:objFunc})--(\ref{equ:solution:benders:sub:cons}) takes as an input, $\bar{x}$, which is fixed equal to a solution of~(\ref{equ:solution:benders:master:objFunc})--(\ref{equ:solution:benders:master:origCons}) and determines resultant values of $u$, $\mu$, and $\lambda$ that are optimal in the lower level of~(\ref{equ:solution:combine:final:objFunc})--(\ref{equ:solution:combine:final:proactive:binary}). Constraints~(\ref{equ:solution:benders:sub:cons}) are the restrictions on the sub-problem variables that are in~(\ref{equ:solution:combine:final:objFunc})--(\ref{equ:solution:combine:final:proactive:binary}).

The sub-problem is feasible for any choice of $\bar{x}$. This feasibility is because $u$ can be chosen to satisfy~(\ref{equ:solution:combine:final:nature:local})--(\ref{equ:solution:combine:final:nature:global}) and $\mu$ and $\lambda$ can be made arbitrarily small to satisfy~(\ref{equ:solution:combine:final:dualCons:struct})--(\ref{equ:solution:combine:final:dualCons:nonNeg:mu}). Thus, solving~(\ref{equ:solution:benders:sub:objFunc})--(\ref{equ:solution:benders:sub:cons}) yields two possible outcomes. If $\Phi(\bar{x}) \geq \theta$, the values of $\bar{x}$, $u$, $\lambda$, and $\mu$ that are obtained from solving the master and sub-problem are optimal in~(\ref{equ:solution:combine:final:objFunc})--(\ref{equ:solution:combine:final:proactive:binary}). Otherwise, if $\Phi(\bar{x}) < \theta$, $\bar{x}$ may be suboptimal in~(\ref{equ:solution:combine:final:objFunc})--(\ref{equ:solution:combine:final:proactive:binary}) because the associated value of $\theta$ underestimates the resultant value of~(\ref{equ:solution:combine:final:objFunc}). In this case, an optimality cut of the form:
\begin{equation}
\label{equ:solution:benders:optCut}
	\theta \geq \Phi(\bar{x}) + \phi^\top (x - \bar{x});
\end{equation}
is added to the set, $\Theta(x)$. $\phi \in \partial \Phi(\bar{x})$ is a subgradient of $\Phi(x)$ at $\bar{x}$. Because~(\ref{equ:solution:benders:sub:objFunc})--(\ref{equ:solution:benders:sub:cons}) is linear in $u$, $\lambda$, and $\mu$, a subgradient can be obtained from the basis matrix and dual-variable values that are found by solving~(\ref{equ:solution:benders:sub:objFunc})--(\ref{equ:solution:benders:sub:cons}). Optimality cut~(\ref{equ:solution:benders:optCut}) uses the sensitivity information that is obtained from solving~(\ref{equ:solution:benders:sub:objFunc})--(\ref{equ:solution:benders:sub:cons}) to improve the estimate of $\theta$ that is given by $\Theta(x)$.

Algorithm~\ref{alg:solution:benders} is pseudocode that details the steps of our implementation of Benders's decomposition. Line~\ref{alg:solution:benders:input} takes $\mathcal{T}$ and $\epsilon$ as inputs. Line~\ref{alg:solution:benders:init} initializes by setting the iteration counter, $t$, to zero, the set, $\Theta(x)$, as empty, and upper and lower bounds of~(\ref{equ:solution:combine:final:objFunc}) to $+\infty$ and $-\infty$, respectively.

\begin{algorithm}
\caption{Benders's Decomposition}
\label{alg:solution:benders}
\begin{algorithmic}[1]
	\STATE \textbf{input:} $\mathcal{T}$, $\epsilon$
\label{alg:solution:benders:input}
	\STATE \textbf{initialize:} $t \gets 0$, $\Theta(x) \gets \emptyset$, $\Phi^+ \gets +\infty$, $\Phi^- \gets -\infty$
\label{alg:solution:benders:init}
	\WHILE{$\Phi^+ - \Phi^- > \epsilon$ \textbf{and} $t < \mathcal{T}$}
\label{alg:solution:benders:loop:start}
		\STATE $(\theta^t,x^t) \gets \argmin \textrm{(\ref{equ:solution:benders:master:objFunc}) s.t. (\ref{equ:solution:benders:master:optimality}), (\ref{equ:solution:benders:master:origCons})}$
\label{alg:solution:benders:master}
		\STATE $(u^t,\lambda^t,\mu^t) \gets \argmax \textrm{(\ref{equ:solution:benders:sub:objFunc}) s.t. (\ref{equ:solution:benders:sub:cons})}$
\label{alg:solution:benders:sub}
		\STATE $\Phi^+ \gets \min\{\Phi^+,\theta^t\}$
\label{alg:solution:benders:update:bound:up}
		\STATE $\Phi^- \gets \max\{\Phi^-,\Phi(x^t)\}$
\label{alg:solution:benders:update:bound:low}
		\STATE $\phi^t \gets \partial \Phi(x^t)$
\label{alg:solution:benders:update:subgradient}
		\STATE $\Theta(x) \gets \Theta(x) \cup \left\{
			\theta \geq \Phi(x^t) + {\phi^t}^\top (x - x^t)
		\right\}$
\label{alg:solution:benders:update:optCut}
		\STATE $t \gets t+1$
\label{alg:solution:benders:update:iter}
	\ENDWHILE
\label{alg:solution:benders:loop:end}
\end{algorithmic}
\end{algorithm}

Lines~\ref{alg:solution:benders:loop:start}--\ref{alg:solution:benders:loop:end} are the iterative loop. Lines~\ref{alg:solution:benders:master} and~\ref{alg:solution:benders:sub} solve the master problem for $x^t$ which is used to solve the sub-problem. Lines~\ref{alg:solution:benders:update:bound:up} and~\ref{alg:solution:benders:update:bound:low} use these solutions to update the upper and lower bounds on~(\ref{equ:solution:combine:final:objFunc}), respectively. Lines~\ref{alg:solution:benders:update:subgradient} and~\ref{alg:solution:benders:update:optCut} obtain a subgradient of $\Phi(x^t)$ which is used to separate a new optimality cut. Line~\ref{alg:solution:benders:update:iter} updates the iteration count.

Algorithm~\ref{alg:solution:benders} has a finite-convergence guarantee for any non-negative value of $\epsilon$. The guarantee stems from~(\ref{equ:solution:benders:sub:objFunc})--(\ref{equ:solution:benders:sub:cons}) being a linear problem, which guarantees that for any $\bar{x}$, (\ref{equ:solution:benders:sub:objFunc})--(\ref{equ:solution:benders:sub:cons}) has an optimal extreme point of~(\ref{equ:solution:benders:sub:cons}). Because~(\ref{equ:solution:benders:sub:cons}) has a finite number of extreme points, a solution of~(\ref{equ:solution:benders:master:objFunc})--(\ref{equ:solution:benders:master:origCons}) will give $x$ that is optimal in~(\ref{equ:solution:combine:final:objFunc})--(\ref{equ:solution:combine:final:proactive:binary}) once $\Theta(x)$ includes optimality cuts that are generated using all of the extreme points of~(\ref{equ:solution:benders:sub:cons}).

\section{Example}
\label{sec:example}
This section summarizes the results of a synthetic example.

\subsection{Data Simulation}
\label{sec:example:data}
Our example has $n = 10$~regions, $B = 1000$, $C = 1$, and $\forall i = 1,\dots,n$, $b_i$ is sampled randomly from a uniform distribution between $100$ and $1000$ and $c_i = 1$.

Defining $\{\hat{L}_i(w), \hat{T}_i(w)\}_{i=0}^n$, $\hat{u}$, and $\bar{u}$ requires that we simulate a set of historical observations. We do this by assuming that outages at each region follow independent susceptible-infected-recovered dynamics \cite{WeiJiGalvanCouvillonOrellanaMomoh2016,KosmaNikolentzosPanagopoulosSteyaertVazirgiannis2023,ZhuYaoXieQiuQiuWu2025,ChenFiorettoQiuZhu2025}. Specifically, $\forall i = 1,\dots,n$, we let $\nu_i$ denote region~$i$'s total population and assume that $\nu_i = \Upsilon_i(\tau) + \Gamma_i(\tau) + \Xi_i(\tau)$, where $\forall \tau$, $\Upsilon_i(\tau)$, $\Gamma_i(\tau)$, and $\Xi_i(\tau)$ denote, respectively, the number of unaffected, disrupted, and recovered region-$i$ customers during time~$\tau$. For all $i = 1,\dots,n$ and $\tau$, the evolution of the outage process, which results in customers moving between being unaffected, disrupted, and recovered, is governed by:
\begin{gather}
\label{equ:example:data:evolution:unaffect}
	\frac{d \Upsilon_i(\tau)}{d \tau} = -\beta_i \frac{\Upsilon_i(\tau) \Gamma_i(\tau)}{\nu_i};\\
\label{equ:example:data:evolution:disrupt}
	\frac{d \Gamma_i(\tau)}{d \tau} = \beta_i \frac{\Upsilon_i(\tau) \Gamma_i(\tau)}{\nu_i} - \rho_i \Gamma_i(\tau);
\end{gather}
and:
\begin{equation}
\label{equ:example:data:evolution:recover}
	\frac{d \Xi_i(\tau)}{d \tau} = \rho_i \Gamma_i(\tau).
\end{equation}
For all $i = 1,\dots,n$, we have that the disruption rate is:
\begin{multline*}
	\beta_i = 0.3 + 0.01 \cdot \left(w_{1,i} - 20\right) + 0.005 \cdot \left(w_{2,i} - 10\right)\\
		- 0.005 \cdot \left(w_{3,i} - 0.5\right);
\end{multline*}
where $w_{1,i}$, $w_{2,i}$, and $w_{3,i}$ are region~$i$'s temperature ($^\circ$~C), wind speed (m$/$s), and relative humidity (\%), and we have recovery rate, $\rho_i = 0.1$. For all $i = 1,\dots,n$, we compute the spatially correlated weather as:
\begin{gather}
\label{equ:example:data:weather:temperature}
	w_{1,i} = 20 + 5 \sin(2 \pi q_i) \cos(2 \pi r_i);\\
\label{equ:example:data:weather:wind}
	w_{2,i} = 10 + 3 \cos(2 \pi q_i) \sin(2 \pi r_i);
\end{gather}
and:
\begin{equation}
\label{equ:example:data:weather:humidity}
	w_{3,i} = 0.5 + 0.1 \sin(2 \pi q_i) \cos(2 \pi r_i);
\end{equation}
where $(q_i,r_i)$ are the coordinates (normalized to the unit square) of region~$i$'s spatial center. Differential equations~(\ref{equ:example:data:evolution:unaffect})--(\ref{equ:example:data:evolution:recover}) are solved using Euler method with fixed time step, $\Delta \tau = 0.1$, with the boundary condition, $\Upsilon_i(0) = \nu_i$, until $\Gamma_i(\tau) = 0$, $\forall i = 1,\dots,n$, (\ie, all disruptions are resolved).

Given the set, $w_1$, $w_2$, and $w_3$, of weather observations that are produced by~(\ref{equ:example:data:weather:temperature})--(\ref{equ:example:data:weather:humidity}) and the resultant $\{\Gamma_i(\tau)\}_{i=1}^n$, region-$i$ outages are simulated as:
\begin{equation}
\label{equ:example:data:outage}
	u_i = \frac{1}{\nu_i} \int_0^{+\infty} \Gamma_i(\tau) d \tau + \ell_i;
\end{equation}
where $\ell_i$ is Gaussian noise with zero mean and variance equal to $\chi^2/\nu_i^2$ and the integral is approximated numerically using fixed time step, $\Delta \tau$. $\chi$ is a global noise parameter, which we assume is $0.1$. Having $\nu_i$ in the denominator of the variance captures the intuition and real-world observation that more (densely) populated regions have lower variance in outage observations compared to less-populated regions. We use~(\ref{equ:example:data:outage}) to generate $200$~random samples of historical observations. $160$~samples define the set, $\mathcal{M}$. The values of $\bar{u}$ that correspond to the remaining $40$~samples are used to evaluate solution quality. The results that are presented in Section~\ref{sec:example:result} are averages across this second set of $40$~samples. For all $i = 1,\dots,n$, $h_i$ is proportional to $\nu_i$ (\ie, disruption cost is increasing in the number of customers who are impacted).

All optimization models are programmed and solved with Python~$3.11.1$ and Gurobi~$11.0.1$, respectively, on a computer with a $10$-core CPU and $16$~GB of memory.

\subsection{Results}
\label{sec:example:result}
Table~\ref{tab:example:result:bilevel} summarizes the performance of our proposed tri-level model and of Proactive-Only and Co-Optimized Planning based on~(\ref{equ:model:benchmark:bilevel}). Although all four methods of forecasting $\hat{u}$ are used as inputs to Proactive-Only Planning, a single set of identical values are reported. The values are identical due to Proactive-Only Planning making decisions based on solely on the relative rankings of the values of $\{h_i \hat{u}_i\}_{i=1}^n$. These values differ between the four forecast-generation methods, but their relative rankings are identical, which gives the same actions regardless of the forecast that is employed.

\begin{table}[hbt]
\centering
\caption{Average (Across $40$~Samples) Value of~(\ref{equ:model:benchmark:bilevel}) for Example from Section~\ref{sec:example} (\$~Thousand)}
\label{tab:example:result:bilevel}
\begin{tabular}{ll lll}
	\hline
	\hline
	\noalign{\smallskip}
	\multicolumn{2}{l}{Planning and}&		\multicolumn{3}{l}{Definition of $\bar{\mathcal{U}}$}\\
							\cline{3-5}
	\multicolumn{2}{l}{Forecasting Methods}&	Local Only&	Global Only&	(\ref{equ:model:predict:cons:nature})\\
	\noalign{\smallskip}
	\hline
	\noalign{\smallskip}
	\multicolumn{2}{l}{\textit{Proactive-Only}}&	$551.5$&	$569.4$&	$520.6$\\
	\multicolumn{2}{l}{\textit{Co-Optimized}}\\
	&	Empirical Average&			$530.1$&	$531.8$&	$495.2$\\
	&	Empirical Conservative&			$551.5$&	$569.4$&	$520.6$\\
	&	Conformal Average&			$529.6$&	$569.4$&	$495.1$\\
	&	Conformal Conservative&			$529.2$&	$535.0$&	$497.4$\\
	\multicolumn{2}{l}{\textit{Tri-Level}}&		$505.0$&	$531.9$&	$495.0$\\
	\noalign{\smallskip}
	\hline
	\hline
\end{tabular}
\end{table}

Table~\ref{tab:example:result:bilevel} shows that with one exception (Co-Optimized Planning with Empirical Average forecasts and global prediction intervals only), our tri-level model outperforms the other two approaches. Our tri-level model has particular performance benefits in the worst-case of having only local prediction intervals. Table~\ref{tab:example:result:recourse} summarizes the performance of the tri-level model and Co-Optimized Planning based on~(\ref{equ:model:benchmark:recourse}), with the former outperforming the latter. This finding demonstrates that conformal prediction in the middle level of~(\ref{equ:model:optimization:objFunc}) provides calibrated, data-driven uncertainty sets that capture heteroskedastic and region-specific variability, while mitigating the risk of overfitting to individual point estimates. Table~\ref{tab:example:result:recourse} reinforces that the tri-level model improves robustness and performance under extreme and more-typical, less-extreme outage scenarios.

\begin{table}[hbt]
\centering
\caption{Average (Across $40$~Samples) Value of~(\ref{equ:model:benchmark:recourse}) for Example from Section~\ref{sec:example} (\$~Thousand)}
\label{tab:example:result:recourse}
\begin{tabular}{ll lll}
	\hline
	\hline
	\noalign{\smallskip}
	\multicolumn{2}{l}{Planning and}&		\multicolumn{3}{l}{Value of $\bar{u}_i$}\\
							\cline{3-5}
	\multicolumn{2}{l}{Forecasting Methods}&	$\mu_i$&	$\mu_i + \sigma_i$&	$\mu_i + 2 \sigma_i$\\
	\noalign{\smallskip}
	\hline
	\noalign{\smallskip}
	\multicolumn{2}{l}{\textit{Co-Optimized}}\\
	&	Empirical Average&			$496.1$&	$559.7$&		$625.2$\\
	&	Empirical Conservative&			$520.7$&	$567.9$&		$599.1$\\
	&	Conformal Average&			$508.8$&	$573.8$&		$647.6$\\
	&	Conformal Conservative&			$532.5$&	$574.9$&		$644.3$\\
	\multicolumn{2}{l}{\textit{Tri-Level}}&		$487.0$&	$505.0$&		$622.8$\\
	\noalign{\smallskip}
	\hline
	\hline
\end{tabular}
\end{table}

\begin{figure}[hbt]
\centering
\includegraphics[clip,keepaspectratio,width=\linewidth]{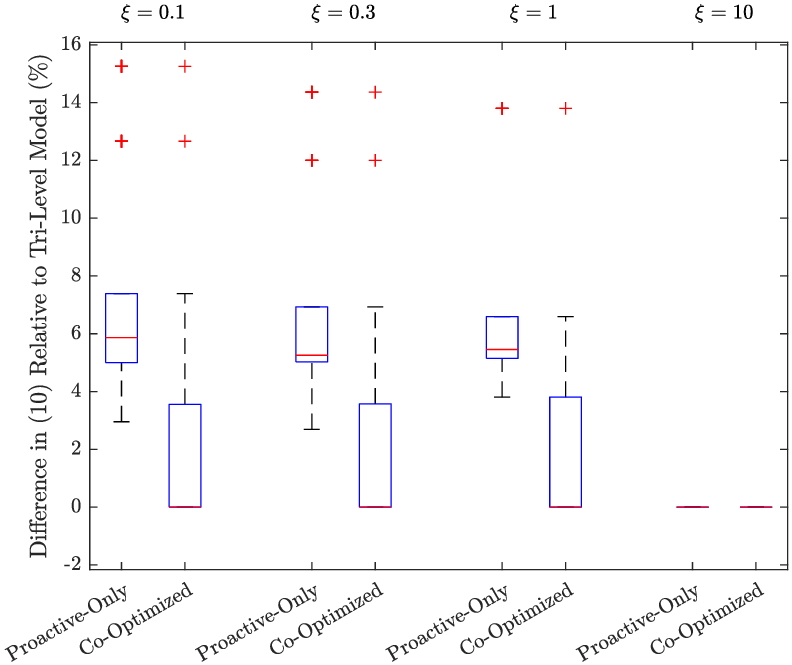}
\vspace{-.2in}
\caption{Difference in~(\ref{equ:model:benchmark:bilevel}) \textit{vis-\`{a}-vis} tri-level model for example in Section~\ref{sec:example} as $\chi$ is varied.}
\label{fig:example_result_sens_noise}
\end{figure}

\begin{figure}[hbt]
\centering
\includegraphics[clip,keepaspectratio,width=\linewidth]{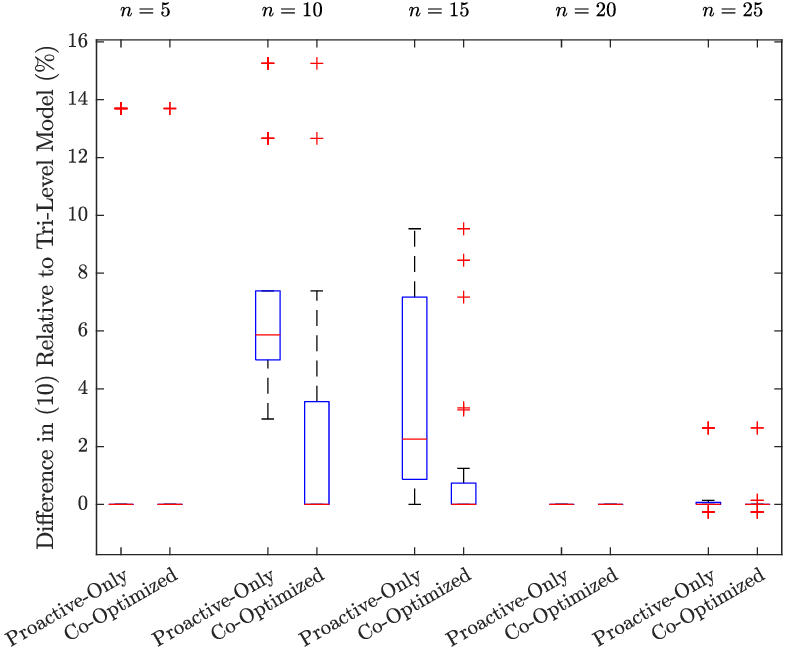}
\vspace{-.2in}
\caption{Difference in~(\ref{equ:model:benchmark:bilevel}) \textit{vis-\`{a}-vis} tri-level model for example in Section~\ref{sec:example} as $n$ is varied.}
\label{fig:example_result_sens_region}
\end{figure}

\begin{figure}[hbt]
\centering
\includegraphics[clip,keepaspectratio,width=\linewidth]{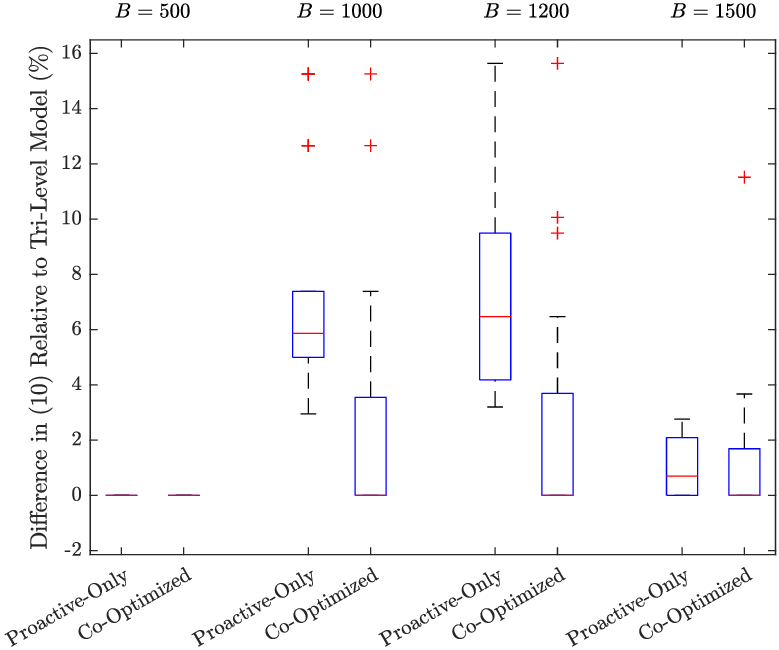}
\vspace{-.2in}
\caption{Difference in~(\ref{equ:model:benchmark:bilevel}) \textit{vis-\`{a}-vis} tri-level model for example in Section~\ref{sec:example} as $B$ is varied.}
\label{fig:example_result_sens_budget_proactive}
\end{figure}

\begin{figure}[hbt]
\centering
\includegraphics[clip,keepaspectratio,width=\linewidth]{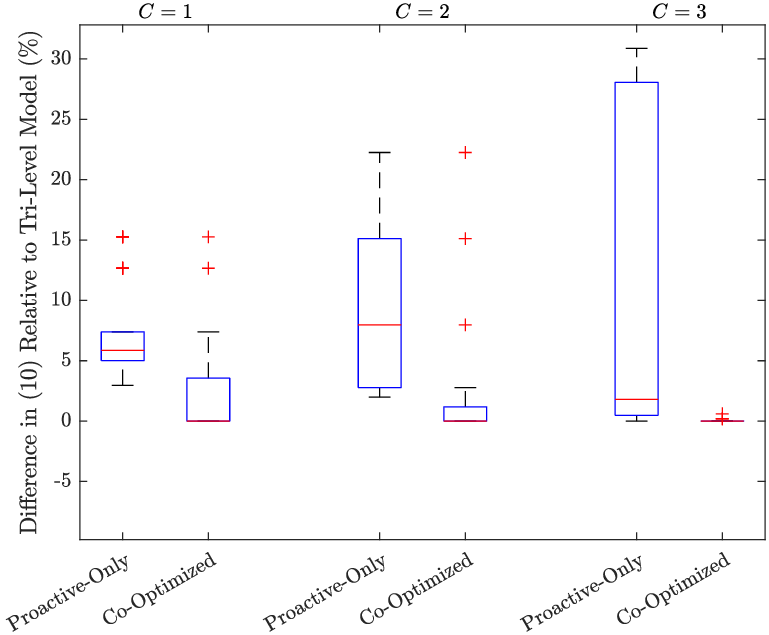}
\vspace{-.2in}
\caption{Difference in~(\ref{equ:model:benchmark:bilevel}) \textit{vis-\`{a}-vis} tri-level model for example in Section~\ref{sec:example} as $C$ is varied.}
\label{fig:example_result_sens_budget_reactive}
\end{figure}

Figs.~\ref{fig:example_result_sens_noise}--\ref{fig:example_result_sens_budget_reactive} are standard box plots that summarize the difference in~(\ref{equ:model:benchmark:bilevel}) between each of Proactive-Only and Co-Optimized Planning (considering the four approaches to generate $\hat{u}$) and our tri-level model with varied input parameters. Fig.~\ref{fig:example_result_sens_noise} shows that as the noise parameter is increased, initially our tri-level model excels (especially under extreme events) due to its use of calibrated uncertainty bounds. This performance is diminished as noise increases further due to the prediction intervals becoming wide and imprecise. Fig.~\ref{fig:example_result_sens_region} shows a similar result for increasing the number of regions---initially the tri-level model excels due to the importance of allocating scarce resources efficiently. However, the value of the tri-level model diminishes as $n$ increases further, due to our holding the proactive and reactive budgets fixed. Fig.~\ref{fig:example_result_sens_budget_proactive} demonstrates similar relative performance of our tri-level model as the proactive budget is increased. Initially, the tri-level model performs well due to its allocating scarce resources efficiently. This advantage diminishes as the budget increases further, due to most regions being protected. Fig.~\ref{fig:example_result_sens_budget_reactive} shows an opposite effect of increasing the reactive budget, which reduces the importance of proactive planning and the relative value of the tri-level model.

\begin{figure}[hbt]
\centering
\includegraphics[clip,keepaspectratio,width=\linewidth]{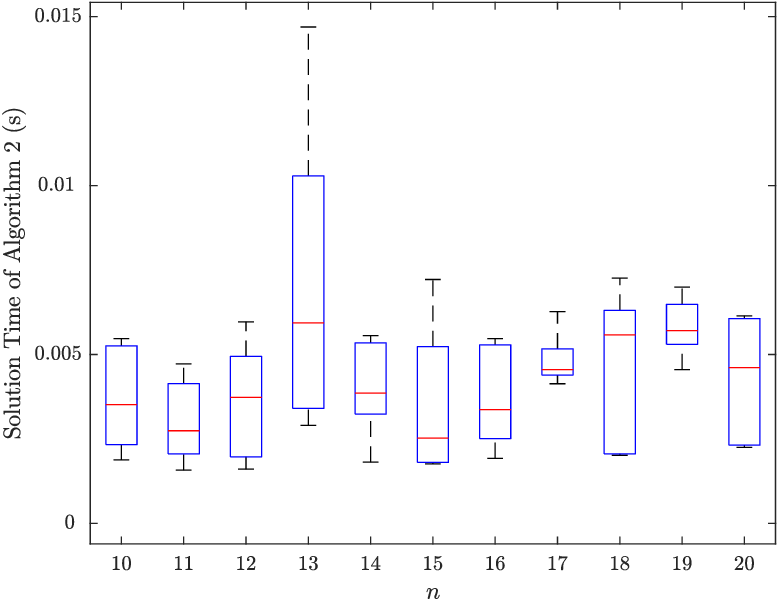}
\vspace{-.2in}
\caption{Total wall-clock time of Algorithm~\ref{alg:solution:benders} for example in Section~\ref{sec:example} as $n$ is varied.}
\label{fig:example_result_scale_time}
\end{figure}

\begin{figure}[hbt]
\centering
\includegraphics[clip,keepaspectratio,width=\linewidth]{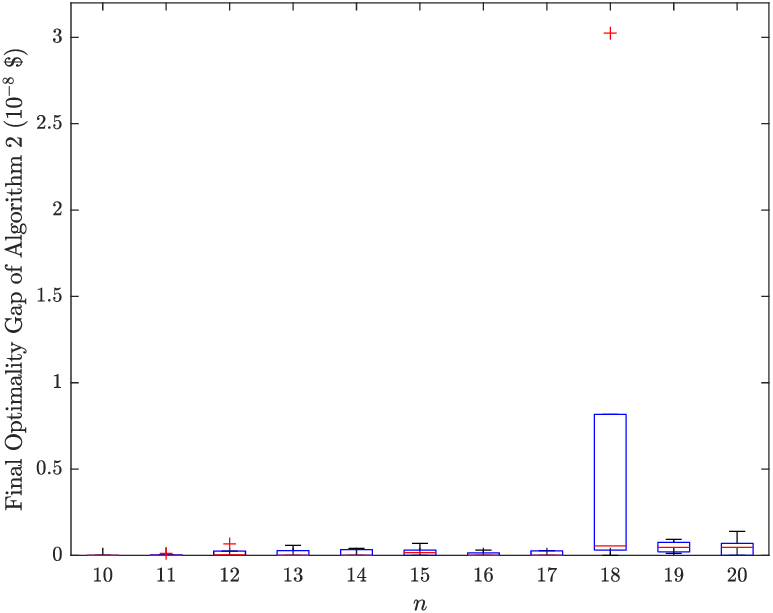}
\caption{Final optimality gap of Algorithm~\ref{alg:solution:benders} for example in Section~\ref{sec:example} as $n$ is varied.}
\label{fig:example_result_scale_opt}
\end{figure}

Figs.~\ref{fig:example_result_scale_time} and~\ref{fig:example_result_scale_opt} examine our model's scalability by summarizing total average (over five of the $40$~samples that are used for solution-quality evaluation) solution time and the optimality gap, $\Phi^+ - \Phi^-$, of the final solution that is found by applying Algorithm~\ref{alg:solution:benders} to instances of our example with different values of $n$.

\section{Case Study}
\label{sec:caseStudy}
This section summarizes the results of a case study.

\subsection{Data}
\label{sec:caseStudy:data}
Our case study is based on three major snowfall events that impacted the northeastern US during $1$--$15$~March,~$2018$. Our focus is Massachusetts, considering the state's $n = 14$~counties as regions. For all $i = 1,\dots,n$, $b_i$, is sampled randomly from a uniform distribution between $100$ and $500$. Case-study results are averaged over five independent samples of these values. We assume $B = 10000$, $C = 2$, and $c_i = 1$, $\forall i = 1,\dots,n$.

The set, $\mathcal{M}$, is generated using weather, socioeconomic, and outage data that correspond to the three $2018$ events. Each event, which occurred during $1$--$3$~March, $6$--$8$~March, and $12$--$15$~March, respectively, is a separate observation. The first two events are used as $\mathcal{M}_\mathrm{train}$ and the other as $\mathcal{M}_\mathrm{cal}$. For each observation, the matrix, $w$, of contextual features consists of weather (\eg, wind speed, temperature, and barometric pressure) and socioeconomic (\eg, median income, age, commute time, and household size; percentage of households receiving supplemental-nutrition assistance; and unemployment, poverty, and college-enrollment rates) data for each county. Weather and socioeconomic data are obtained from National Oceanic and Atmospheric Administration\footnote{\url{https://rapidrefresh.noaa.gov/hrrr/}} and US Census Bureau's $2017$~American Community Survey, respectively. Outage data for the three events are based on reports by the electric utilities to Massachusetts Department of Public Utilities.\footnote{\cf\ Docket numbers $19$-ERP-$08$, $19$-ERP-$09$, and $19$-ERP-$10$.} For all $i = 1,\dots,n$, the value of $h_i$ is proportional to the number of utility customers that are in county~$i$. The case study uses the same computational resources that are used for the example.

\subsection{Results}
Table~\ref{tab:caseStudy:result:bilevel} provides the same information for the case study that Table~\ref{tab:example:result:bilevel} does for the example. Table~\ref{tab:caseStudy:result:bilevel} reinforces the value of our tri-level model in planning for worst-case outcomes. The deterministic planning approaches outperform our tri-level model if outages do not have local-interval constraints. Given geospatial correlation of electricity-system outages, Table~\ref{tab:caseStudy:result:bilevel} illustrates that our tri-level model and conformal predictions provide robustness to the true outage-uncertainty structure.

\begin{table}[hbt]
\centering
\caption{Average (Across Five Samples) Value of~(\ref{equ:model:benchmark:bilevel}) for Case Study from Section~\ref{sec:caseStudy} (\$~Million)}
\label{tab:caseStudy:result:bilevel}
\begin{tabular}{ll lll}
	\hline
	\hline
	\noalign{\smallskip}
	\multicolumn{2}{l}{Solution and}&		\multicolumn{3}{l}{Definition of $\bar{\mathcal{U}}$}\\
							\cline{3-5}
	\multicolumn{2}{l}{Forecasting Methods}&	Local Only&	Global Only&	(\ref{equ:model:predict:cons:nature})\\
	\noalign{\smallskip}
	\hline
	\noalign{\smallskip}
	\multicolumn{2}{l}{\textit{Proactive-Only}}&	$10.45$&	$14.02$&	$8.65$\\
	\multicolumn{2}{l}{\textit{Co-Optimized}}\\
	&	Empirical Average&			$13.03$&	$15.16$&	$10.46$\\
	&	Empirical Conservative&			$11.50$&	$13.72$&	$9.34$\\
	&	Conformal Average&			$11.50$&	$13.72$&	$9.34$\\
	&	Conformal Conservative&			$7.64$&		$27.33$&	$7.00$\\
	\multicolumn{2}{l}{\textit{Tri-Level}}&		$7.86$&		$19.45$&	$6.68$\\
	\noalign{\smallskip}
	\hline
	\hline
\end{tabular}
\end{table}

As Table~\ref{tab:example:result:recourse} does for the example, Table~\ref{tab:caseStudy:result:recourse} reports the value of~(\ref{equ:model:benchmark:recourse}) for the case study. The final three columns of Table~\ref{tab:caseStudy:result:recourse} use different values of $\bar{u}$ than Table~\ref{tab:example:result:recourse} does. The column that is labeled `Third Storm' uses actual outage data for the third storm. The other two columns use actual outage data for the third storm, plus one or two standard errors (as estimated from the first two storms). Table~\ref{tab:caseStudy:result:recourse} shows that our tri-level model provides decisions that are more costly than Co-Optimized Planning. There are two considerations that temper that interpretation of Table~\ref{tab:caseStudy:result:recourse}. First, the costs that are reported in Table~\ref{tab:caseStudy:result:recourse} correspond to a single weather event. Consequently, these values do not reflect the long-term benefit of our tri-level model and conformal predictions. Second, the third storm is milder than the other two. Consequently, the conformal predictions that are used in the tri-level model are calibrated to more extreme events, meaning that the model provides decisions that protect against more extreme events.

\begin{table}[hbt]
\centering
\caption{Average (Across Five Samples) Value of~(\ref{equ:model:benchmark:recourse}) for Case Study from Section~\ref{sec:caseStudy} (\$~Million)}
\label{tab:caseStudy:result:recourse}
\begin{tabular}{ll lll}
	\hline
	\hline
	\noalign{\smallskip}
	\multicolumn{2}{l}{Solution and}&		\multicolumn{3}{l}{Value of $\bar{u}_i$}\\
							\cline{3-5}
	\multicolumn{2}{l}{Forecasting Methods}&	Third Event&	$+\sigma_i$&	$+2 \sigma_i$\\
	\noalign{\smallskip}
	\hline
	\noalign{\smallskip}
	\multicolumn{2}{l}{\textit{Co-Optimized}}\\
	&	Empirical Average&			$6.01$&		$7.26$&		$13.93$\\
	&	Empirical Conservative&			$7.74$&		$7.01$&		$44.73$\\
	&	Conformal Average&			$10.67$&	$8.53$&		$16.02$\\
	&	Conformal Conservative&			$14.99$&	$29.87$&	$44.73$\\
	\multicolumn{2}{l}{\textit{Tri-Level}}&		$14.89$&	$27.71$&	$40.53$\\
	\noalign{\smallskip}
	\hline
	\hline
\end{tabular}
\end{table}

Contrasting proactive actions that are selected by the models yields insights into tradeoffs between outage cost and risk. Based on the outages that are caused by the first two storms, outage risk is estimated to be highest at the western inland area of the state, including Berkshire and Hampshire counties. However, the third storm had greatest impact at the eastern coastal area of the state, including Plymouth and Barnstable counties. On the other hand, the state's population and potential outage cost is concentrated at its eastern counties. Consequently, Proactive-Only and Co-Optimized Planning concentrate proactive actions at the western area of the state, to protect against what is estimated from the first two storms as the area with greatest outage risk. Conversely, our tri-level model allocates proactive actions to more populated regions---balancing between outage cost and risk.

\begin{figure}[hbt]
\centering
\includegraphics[clip,keepaspectratio,width=\linewidth]{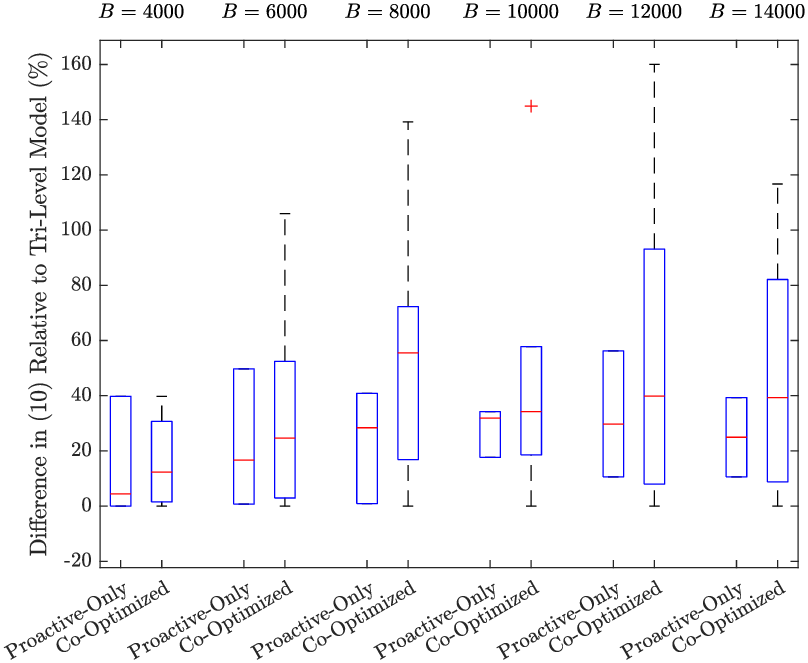}
\vspace{-.2in}
\caption{Difference in~(\ref{equ:model:benchmark:bilevel}) \textit{vis-\`{a}-vis} tri-level model for case study in Section~\ref{sec:caseStudy} as $B$ is varied.}
\label{fig:caseStudy_result_sens_budget_proactive}
\end{figure}

\begin{figure}[hbt]
\centering
\includegraphics[clip,keepaspectratio,width=\linewidth]{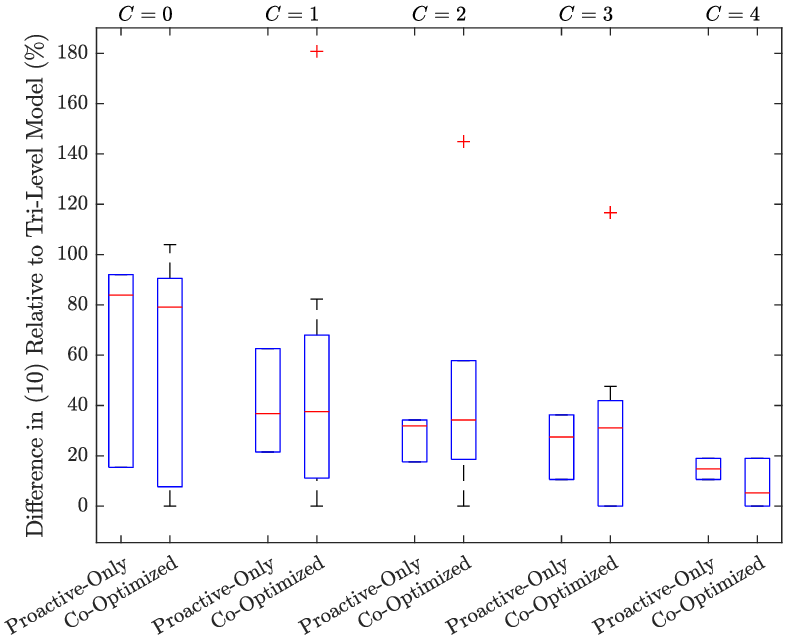}
\vspace{-.2in}
\caption{Difference in~(\ref{equ:model:benchmark:bilevel}) \textit{vis-\`{a}-vis} tri-level model for case study in Section~\ref{sec:caseStudy} as $C$ is varied.}
\label{fig:caseStudy_result_sens_budget_reactive}
\end{figure}

Figs.~\ref{fig:caseStudy_result_sens_budget_proactive} and~\ref{fig:caseStudy_result_sens_budget_reactive} summarize the same metrics for the case study that Figs.~\ref{fig:example_result_sens_budget_proactive} and~\ref{fig:example_result_sens_budget_reactive} do for the example. Fig.~\ref{fig:caseStudy_result_sens_budget_proactive} shows that increasing the proactive budget narrows the performance gap between our tri-level model and Proactive-Only and Co-Optimized Planning. Nonetheless, the tri-level model provides decisions that are more robust to extreme events \textit{vis-\`{a}-vis} the other two planning approaches, which is exemplified by the outliers that are in Fig.~\ref{fig:caseStudy_result_sens_budget_proactive}. Fig.~\ref{fig:caseStudy_result_sens_budget_reactive} shows that increasing the reactive budget decreases the relative benefit of the tri-level model, due to relative reduction in the value of proactive actions (\ie, reactive actions mitigate poor proactive decisions).

\section{Conclusion}
\label{sec:concl}
This paper proposes a model with which to increase electricity-system resilience. By casting the problem as a tri-level optimization that links proactive and reactive actions with uncertain disruptions, we offer a structured model with which to co-optimize long- and short-term resilience actions. Using spatio-temporal conformal prediction, our method constructs adaptive, distribution-free uncertainty sets to address data scarcity and spatial heterogeneity. Although our focus is weather-driven electricity-system disruptions, the model can be adapted to disruption types. Indeed, the middle level of the model could be recast as representing an adversarial attack (\eg, by a human), in which case conformal prediction may be unnecessary.

The model's structure (\ie, tri-level, tri-linear objective-function terms, variable structure, and potentially non-convex uncertainty set) makes it computationally complex. We address these complications to yield a bi-level model that can be solved efficiently with Benders's decomposition. Conformal prediction is used to generate polyhedral uncertainty sets, which are guaranteed to meet the coverage probability that is specified by~(\ref{equ:model:optimization:cons:nature}). Lemma~\ref{lem:solution:continuous} shows that the lowest-level problem can be replaced with its equivalent continuous relaxation, which allows (through the application of strong duality) to convert the tri-level model into an equivalent bi-level formulation. Benders's decomposition solves the bi-level model efficiently.

Numerical experiments using a synthetic example and a real-world case study that is based on a North American winter storm show that the tri-level structure of our model yields solutions that outperform simpler optimization approaches. The benefit of our model is pronounced especially under tight budget constraints. Overall, our results underscore the importance of anticipatory planning, robust uncertainty modeling, and scalable solvers in building resilient power systems.


\begin{thebibliography}{10}
\providecommand{\url}[1]{#1}
\csname url@samestyle\endcsname
\providecommand{\newblock}{\relax}
\providecommand{\bibinfo}[2]{#2}
\providecommand{\BIBentrySTDinterwordspacing}{\spaceskip=0pt\relax}
\providecommand{\BIBentryALTinterwordstretchfactor}{4}
\providecommand{\BIBentryALTinterwordspacing}{\spaceskip=\fontdimen2\font plus
\BIBentryALTinterwordstretchfactor\fontdimen3\font minus
  \fontdimen4\font\relax}
\providecommand{\BIBforeignlanguage}[2]{{%
\expandafter\ifx\csname l@#1\endcsname\relax
\typeout{** WARNING: IEEEtran.bst: No hyphenation pattern has been}%
\typeout{** loaded for the language `#1'. Using the pattern for}%
\typeout{** the default language instead.}%
\else
\language=\csname l@#1\endcsname
\fi
#2}}
\providecommand{\BIBdecl}{\relax}
\BIBdecl

\bibitem{StankovicTomsovic2018}
A.~M. Stankovi\'{c} and K.~L. Tomsovic, ``{The Definition and Quantification of Resilience},'' IEEE PES Industry Technical Support Task Force {PES-TR65}, April 2018.

\bibitem{HeHuangLiuZhang2021}
H.~He, S.~Huang, Y.~Liu, and T.~Zhang, ``{A tri-level optimization model for power grid defense with the consideration of post-allocated DGs against coordinated cyber-physical attacks},'' \emph{International Journal of Electrical Power \& Energy Systems}, vol. 130, p. 106903, September 2021.

\bibitem{LiuShahidehpourLiLiuCaoBie2017}
X.~Liu, M.~Shahidehpour, Z.~Li, X.~Liu, Y.~Cao, and Z.~Bie, ``{Microgrids for Enhancing the Power Grid Resilience in Extreme Conditions},'' \emph{IEEE Transactions on Smart Grid}, vol.~8, pp. 589--597, March 2017.

\bibitem{PanteliTrakasMancarellaHatziargyriou2017}
M.~Panteli, D.~N. Trakas, P.~Mancarella, and N.~D. Hatziargyriou, ``{Power Systems Resilience Assessment: Hardening and Smart Operational Enhancement Strategies},'' \emph{Proceedings of the IEEE}, vol. 105, pp. 1202--1213, July 2017.

\bibitem{StankovicTomsovicDeCaroBraunChowCukalevski2023}
A.~M. Stankovi\'{c}, K.~L. Tomsovic, F.~{De Caro}, M.~Braun, J.~H. Chow, and N.~\v{C}ukalevski, ``{Methods for Analysis and Quantification of Power System Resilience},'' \emph{IEEE Transactions on Power Systems}, vol.~38, pp. 4774--4787, September 2023.

\bibitem{HuangWangChenQiGuo2017}
G.~Huang, J.~Wang, C.~Chen, J.~Qi, and C.~Guo, ``{Integration of Preventive and Emergency Responses for Power Grid Resilience Enhancement},'' \emph{IEEE Transactions on Power Systems}, vol.~32, pp. 4451--4463, November 2017.

\bibitem{LinChenWangBie2019}
Y.~Lin, B.~Chen, J.~Wang, and Z.~Bie, ``{A Combined Repair Crew Dispatch Problem for Resilient Electric and Natural Gas System Considering Reconfiguration and DG Islanding},'' \emph{IEEE Transactions on Power Systems}, vol.~34, pp. 2755--2767, July 2019.

\bibitem{LeiChenLiHou2019}
S.~Lei, C.~Chen, Y.~Li, and Y.~Hou, ``{Resilient Disaster Recovery Logistics of Distribution Systems: Co-Optimize Service Restoration With Repair Crew and Mobile Power Source Dispatch},'' \emph{IEEE Transactions on Smart Grid}, vol.~10, pp. 6187--6202, November 2019.

\bibitem{MohamedChenSuJin2019}
M.~A. Mohamed, T.~Chen, W.~Su, and T.~Jin, ``{Proactive Resilience of Power Systems Against Natural Disasters: A Literature Review},'' \emph{IEEE Access}, vol.~7, pp. 163\,778--163\,795, 2019.

\bibitem{MaChenWang2018}
S.~Ma, B.~Chen, and Z.~Wang, ``{Resilience Enhancement Strategy for Distribution Systems Under Extreme Weather Events},'' \emph{IEEE Transactions on Smart Grid}, vol.~9, pp. 1442--1451, March 2018.

\bibitem{BaoZhangOuyangMiao2019}
S.~Bao, C.~Zhang, M.~Ouyang, and L.~Miao, ``{An integrated tri-level model for enhancing the resilience of facilities against intentional attacks},'' \emph{Annals of Operations Research}, vol. 283, pp. 87--117, December 2019.

\bibitem{WangRousisStrbac2021}
Y.~Wang, A.~O. Rousis, and G.~Strbac, ``{A Three-Level Planning Model for Optimal Sizing of Networked Microgrids Considering a Trade-Off Between Resilience and Cost},'' \emph{IEEE Transactions on Power Systems}, vol.~36, pp. 5657--5669, November 2021.

\bibitem{HinesDobsonRezaei2017}
P.~D.~H. Hines, I.~Dobson, and P.~Rezaei, ``{Cascading Power Outages Propagate Locally in an Influence Graph That is Not the Actual Grid Topology},'' \emph{IEEE Transactions on Power Systems}, vol.~32, pp. 958--967, March 2017.

\bibitem{VovkGammermanShafer2022}
V.~Vovk, A.~Gammerman, and G.~Shafer, \emph{{Algorithmic Learning in a Random World}}, 2nd~ed.\hskip 1em plus 0.5em minus 0.4em\relax Cham, Switzerland: Springer Nature Switzerland AG, July 2022.

\bibitem{ZengZhao2013}
B.~Zeng and L.~Zhao, ``{Solving two-stage robust optimization problems using a column-and-constraint generation method},'' \emph{Operations Research Letters}, vol.~41, pp. 457--461, September 2013.

\bibitem{PatelRayanTewari2024}
Y.~Patel, S.~Rayan, and A.~Tewari, ``{Conformal Contextual Robust Optimization},'' \emph{Proceedings of Machine Learning Research}, vol. 238, pp. 2485--2493, 2-4 May 2024.

\bibitem{BertsimasBrownCaramanis2011}
D.~Bertsimas, D.~B. Brown, and C.~Caramanis, ``{Theory and Applications of Robust Optimization},'' \emph{SIAM Review}, vol.~53, pp. 464--501, 2011.

\bibitem{Benders1962}
J.~F. Benders, ``{Partitioning procedures for solving mixed-variables programming problems},'' \emph{Numerische Mathematik}, vol.~4, pp. 238--252, December 1962.

\bibitem{Jabr2013}
R.~A. Jabr, ``{Robust Transmission Network Expansion Planning With Uncertain Renewable Generation and Loads},'' \emph{IEEE Transactions on Power Systems}, vol.~28, pp. 4558--4567, November 2013.

\bibitem{Ben-TalGoryashkoGuslitzerNemirovski2004}
A.~Ben-Tal, A.~Goryashko, E.~Guslitzer, and A.~Nemirovski, ``{Adjustable robust solutions of uncertain linear programs},'' \emph{Mathematical Programming}, vol.~99, pp. 351--376, March 2004.

\bibitem{BertsimasNaStellatoWang2025}
D.~Bertsimas, L.~Na, B.~Stellato, and I.~Wang, ``{The Benefit of Uncertainty Coupling in Robust and Adaptive Robust Optimization},'' \emph{INFORMS Journal on Optimization}, vol.~7, pp. 105--141, Spring 2025.

\bibitem{BertsimasLitvinovSunZhaoZheng2013}
D.~Bertsimas, E.~Litvinov, X.~A. Sun, J.~Zhao, and T.~Zheng, ``{Adaptive Robust Optimization for the Security Constrained Unit Commitment Problem},'' \emph{IEEE Transactions on Power Systems}, vol.~28, pp. 52--63, February 2013.

\bibitem{Zeng2020}
B.~Zeng, ``{A Practical Scheme to Compute the Pessimistic Bilevel Optimization Problem},'' \emph{INFORMS Journal on Computing}, vol.~32, pp. 1128--1142, Fall 2020.

\bibitem{ZengDongSioshansiXuZeng2020}
B.~Zeng, H.~Dong, R.~Sioshansi, F.~Xu, and M.~Zeng, ``{Bi-Level Robust Optimization of Electric Vehicle Charging Stations with Distributed Energy Resources},'' \emph{IEEE Transactions on Industrial Applications}, vol.~56, pp. 5836--5847, September-October 2020.

\bibitem{NohadaniSharma2018}
O.~Nohadani and K.~Sharma, ``{Optimization under Decision-Dependent Uncertainty},'' \emph{SIAM Journal on Optimization}, vol.~28, pp. 1773--1795, 2018.

\bibitem{BrownCarlyleSalmeronWood2006}
G.~Brown, M.~Carlyle, J.~Salmer\'{o}n, and K.~Wood, ``{Defending Critical Infrastructure},'' \emph{Interfaces}, vol.~36, pp. 530--544, November-December 2006.

\bibitem{EskandarpourKhodaei2017}
R.~Eskandarpour and A.~Khodaei, ``{Machine Learning Based Power Grid Outage Prediction in Response to Extreme Events},'' \emph{IEEE Transactions on Power Systems}, vol.~32, pp. 3315--3316, July 2017.

\bibitem{LiWangGuoZhang2024}
F.~Li, D.~Wang, H.~Guo, and J.~Zhang, ``{Distributionally Robust Optimization for integrated energy system accounting for refinement utilization of hydrogen and ladder-type carbon trading mechanism},'' \emph{Applied Energy}, vol. 367, p. 123391, 1 August 2024.

\bibitem{PapadopoulosGammermanVovk2008}
H.~Papadopoulos, A.~Gammerman, and V.~Vovk, ``{Normalized nonconformity measures for regression Conformal Prediction},'' in \emph{Proceedings of the 26th IASTED International Conference on Artificial Intelligence and Applications}.\hskip 1em plus 0.5em minus 0.4em\relax Innsbruck, Austria: International Association of Science and Technology for Development, 11-13 February 2008, pp. 64--69.

\bibitem{Bellotti2020}
A.~Bellotti, ``{Constructing normalized nonconformity measures based on maximizing predictive efficiency},'' \emph{Proceedings of Machine Learning Research}, vol. 128, pp. 41--54, 9-11 September 2020.

\bibitem{ShaferVovk2008}
G.~Shafer and V.~Vovk, ``{A Tutorial on Conformal Prediction},'' \emph{Journal of Machine Learning Research}, vol.~9, pp. 371--421, 2008.

\bibitem{Wolsey1998}
L.~A. Wolsey, \emph{{Integer Programming}}.\hskip 1em plus 0.5em minus 0.4em\relax New York, New York: Wiley-Interscience, 1998.

\bibitem{SioshansiConejo2017}
R.~Sioshansi and A.~J. Conejo, \emph{{Optimization in Engineering: Models and Algorithms}}, ser. Springer Optimization and Its Applications.\hskip 1em plus 0.5em minus 0.4em\relax Gewerbestra{\ss}e 11, 6330 Cham, Switzerland: Springer Nature, 2017, vol. 120.

\bibitem{WeiJiGalvanCouvillonOrellanaMomoh2016}
Y.~Wei, C.~Ji, F.~Galvan, S.~Couvillon, G.~Orellana, and J.~Momoh, ``{Non-Stationary Random Process for Large-Scale Failure and Recovery of Power Distribution},'' \emph{Applied Mathematics}, vol.~7, pp. 233--249, February 2016.

\bibitem{KosmaNikolentzosPanagopoulosSteyaertVazirgiannis2023}
C.~Kosma, G.~Nikolentzos, G.~Panagopoulos, J.-M. Steyaert, and M.~Vazirgiannis, ``{Neural Ordinary Differential Equations for Modeling Epidemic Spreading},'' \emph{Transactions on Machine Learning Research}, 19 August 2023.

\bibitem{ZhuYaoXieQiuQiuWu2025}
S.~Zhu, R.~Yao, Y.~Xie, F.~Qiu, Y.~Qiu, and X.~Wu, ``{Quantifying Grid Resilience Against Extreme Weather Using Large-Scale Customer Power Outage Data},'' \emph{INFORMS Journal on Data Science}, 2025, forthcoming.

\bibitem{ChenFiorettoQiuZhu2025}
S.~Chen, F.~Fioretto, F.~Qiu, and S.~Zhu, ``{Global-Decision-Focused Neural ODEs for Proactive Grid Resilience Management},'' arXiv, 2025, 2502.18321.
\end{thebibliography}
\end{document}